\documentclass{article}

\usepackage{iclr2027_conference,times}
\usepackage[utf8]{inputenc}
\usepackage{amsmath,amssymb,mathtools}
\usepackage{graphicx}
\usepackage{booktabs}
\usepackage{placeins}
\usepackage{hyperref}
\usepackage{url}

\DeclareUnicodeCharacter{2013}{\textendash}

\title{Ask Self, Ask Others: Relation Is All You Need}
\author{%
\begin{tabular}{c@{\quad}c@{\quad}c}
\begin{tabular}[t]{c}
\textbf{Yuting Ge}$^{1,*}$\\[-1pt]
{\normalfont\scriptsize \textsuperscript{1}Department of Electrical Engineering}\\
{\normalfont\scriptsize City University of Hong Kong}\\
{\normalfont\scriptsize\texttt{yutingge3-c@my.cityu.edu.hk}}
\end{tabular}
&
\begin{tabular}[t]{c}
\textbf{Yang Pengju}$^{2,\dagger}$\\[-1pt]
{\normalfont\scriptsize \textsuperscript{2}Jilin University}\\
{\normalfont\scriptsize\texttt{yangda1223@outlook.com}}
\end{tabular}
&
\begin{tabular}[t]{c}
\textbf{Mingkai Nie}$^{3,\dagger}$\\[-1pt]
{\normalfont\scriptsize \textsuperscript{3}National University of Singapore}\\
{\normalfont\scriptsize\texttt{mingkai.nie@u.nus.edu}}
\end{tabular}%
\end{tabular}%
}

\begin{document}

\maketitle

\begin{center}
{\normalfont\small $^{\dagger}$ These authors contributed equally.\quad
$^*$ Corresponding author.}
\end{center}

\begin{abstract}
Attention dominates token mixing, but it collapses relation formation and flow allocation into a single score-to-flow step. We introduce Relation, which separates them by first organizing pairwise evidence into explicit Self and Exchange relations and deriving information flow afterward. Relation first decides whether a token should rely on itself or draw from its history, and if it draws from history, where to look. This relational organization gives rise to Full Relation, FlashRelation, Linear Relation, and Hybrid Relation. Across matched decoder-only models, Full Relation achieves lower mean final-validation NLL than MHA and reaches the paired MHA final training loss with 4.5\text{--}7.3\% fewer tokens. Structural diagnostics further show that Relation learns a distinct depth organization: the first layer acts as a current-token anchor and a high-rank router, while later layers shift strongly toward history. In a fixed-context reference benchmark, FlashRelation is \(4.17\text{--}5.28\times\) faster than the materialized Full Relation implementation. Across scale-matched production workloads, it reaches \(89.7\text{--}92.9\%\) of PyTorch FlashAttention throughput while executing the exact Full Relation operator. Hybrid Relation demonstrates that Full and Linear Relation layers can be composed within a single decoder. These results support a relation-first view of token mixing: ask Self, ask Others, then let Flow follow Relation.
\end{abstract}


\section{Introduction}

In scaled dot-product Attention, each head maps the interaction between two projected token states to a scalar score \citep{vaswani2017attention},
\[
U_{ij}=\frac{q_i^\top k_j}{\sqrt{d_h}},
\qquad
F_i=\operatorname{Softmax}(U_i).
\]
Token interactions in language are heterogeneous, but Attention reduces each pairwise interaction to a single scalar. The scalar carries too much: the same \(U_{ij}\) must encode whatever relational structure matters and directly determine information flow. Relation formation and flow allocation are therefore compressed into a single \(U\!\rightarrow\!F\) step.

Attention itself was not always primary. It first appeared as an auxiliary mechanism within recurrent encoder--decoder models before becoming the organizing primitive of the Transformer \citep{bahdanau2015neural,vaswani2017attention}. Relation-aware Attention places relation in a similarly subordinate position. \citet{shaw2018relative} augment self-attention with pairwise relation representations, but relation is still inside Attention. Attention remains primary: it incorporates relation into its own computation and still determines the flow. We reverse this hierarchy: Flow follows Relation.

\begin{figure}[t]
\centering
\includegraphics[width=\linewidth]{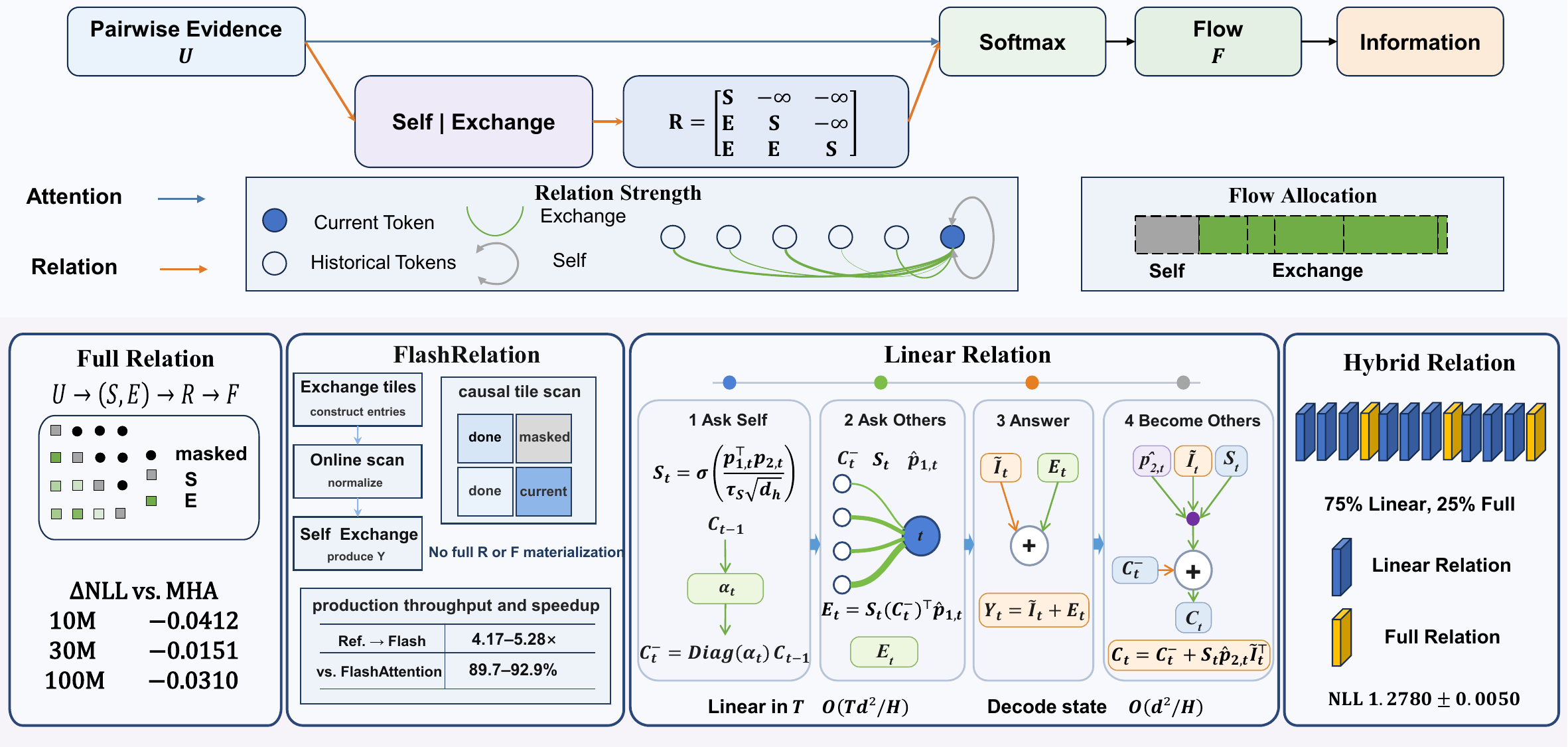}
\caption{Overview of Relation.}
\label{fig:relation-overview}
\end{figure}

We begin by partitioning each causal Relation row into two minimal structural roles: Self and Exchange. Self describes a token's relation with itself, while Exchange describes its relation with other tokens. Pairwise evidence is mapped into these two roles before normalization, and the resulting entries are assembled into a causal Relation matrix \(R\). Normalized information flow \(F\) is derived only afterward. This defines the basic Self--Exchange Relation (SER) operator, which extends naturally to Multi-Head Relation (MHR). This explicit organization also exposes further forms of Relation in practice: Full Relation retains explicit token-wise history, FlashRelation factorizes it for tiled execution, Linear Relation compresses historical Others into a recurrent state, and Hybrid Relation combines Full and Linear layers.

We evaluate Relation in matched decoder-only models at approximately 10M, 30M, and 100M parameters. Full Relation improves mean final-validation NLL at all three scales. Structural ablations examine its behavior, while FlashRelation, Linear Relation, and Hybrid Relation demonstrate practical implementations of the framework (Figure 1). Our main contributions are:

\begin{itemize}
\item We introduce Self--Exchange Relation (SER) and Multi-Head Relation (MHR) as Relation operators for token mixing.
\item We characterize Self--Exchange flow and derive its exact factorization into Self mass, aggregate Exchange mass, and allocation within historical Exchange. Matched structural diagnostics further show that Relation learns an internal organization distinct from Attention.
\item Building on Relation's distinct relational organization, we develop Relation-specific practical forms, including FlashRelation, Linear Relation, and Hybrid Relation, to make the framework practical across exact fused, recurrent, and hybrid execution.
\end{itemize}

\section{Background}

Scaled dot-product Attention maps pairwise compatibility scores directly to normalized token-to-token flow \citep{vaswani2017attention}. Several lines modify this pipeline without changing its organizing primitive. Relative-position methods add pairwise representations or structured biases to Attention scores \citep{shaw2018relative,raffel2020t5,press2022alibi,chi2022kerple}, Sigmoid self-attention changes normalization \citep{ramapuram2024sigmoid}, Talking-Heads Attention mixes across the head dimension around normalization \citep{shazeer2020talkingheads}, and GAU combines attention with gating \citep{hua2022transformerquality}. These methods alter how Attention scores or flows are formed, but Attention remains the organizing primitive. Relation instead makes pairwise relation formation an explicit stage before normalization.

Beyond Attention, alternative token-mixing operators explore different ways to organize information exchange across a sequence. FNet uses Fourier mixing \citep{leethorp2022fnet}, PoNet uses multi-granularity pooling \citep{tan2022ponet}, and HyperMixer constructs token-mixing MLPs \citep{mai2023hypermixer}. These approaches organize token interaction through different mixing operators, whereas Relation organizes token mixing around explicit pairwise relations.

A third line changes how token mixing is executed or how history is represented. FlashAttention and FlashAttention-2 provide tiled, IO-aware execution of exact softmax Attention \citep{dao2022flashattention,dao2023flashattention2}. Linear and recurrent methods replace explicit quadratic token-token interaction with associative or recurrent states \citep{katharopoulos2020linear,sun2023retnet,yang2024gla,yang2025gatedelta}. KDA further develops recurrent-state token mixing, and Kimi Linear combines recurrent KDA with full-attention MLA layers in a hybrid architecture \citep{kimiLinear2025}. Relation likewise admits efficient and recurrent realizations, while preserving the same principle: Flow follows Relation.

\section{Relation}

We first introduce Self–Exchange Relation (SER), the basic Relation operator. We then extend SER to Multi-Head Relation (MHR), as illustrated in Figure 2.

\setcounter{figure}{1}
\begin{figure}[!ht]
\centering
\includegraphics[width=\linewidth]{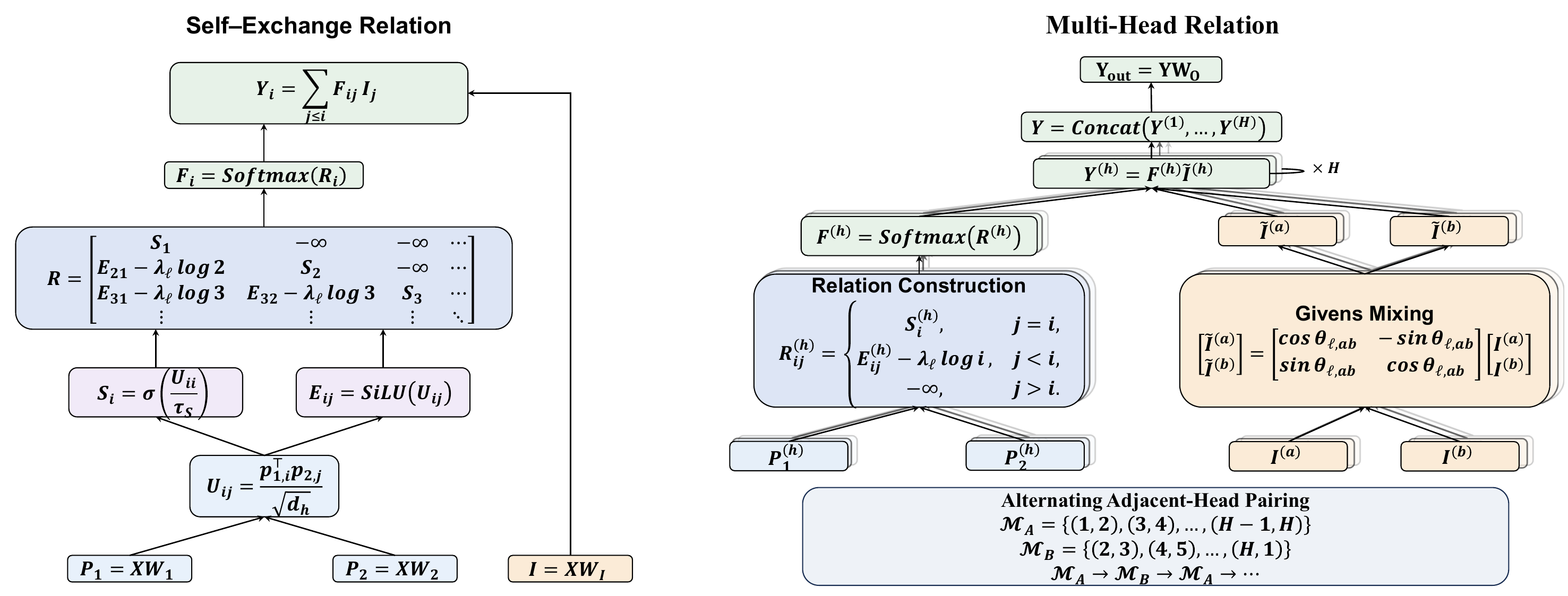}
\caption{Self–Exchange Relation and Multi-Head Relation.}
\label{fig:self-exchange}
\end{figure}
\FloatBarrier

\subsection{Self–Exchange Relation}

Given an input sequence $X$, we project it into two relation spaces and an information space,

\[
P_1=XW_1,\qquad P_2=XW_2,\qquad I=XW_I.
\]

\begin{equation}
U_{ij}=\frac{p_{1,i}^{\top}p_{2,j}}{\sqrt{d_h}}.
\label{eq:relation-projections}
\end{equation}

The head-wise \(P_1\) and \(P_2\) entering \(U\) use RoPE \citep{su2024roformer}, while \(I\) remains unrotated. We use two basic relational roles: Self (\textit{ask self}), which measures the current token itself, and Exchange (\textit{ask others}), which measures other tokens. We map pairwise evidence into these roles as

\begin{equation}
\begin{gathered}
S_i=\sigma\left(\frac{U_{ii}}{\tau_S}\right),
\qquad E_{ij}=\operatorname{SiLU}\left(U_{ij}\right),\\[0.35em]
\begin{gathered}
R_{ij}=
\begin{cases}
S_i, & j=i,\\
E_{ij}-\lambda_\ell\log i, & j<i,\\
-\infty, & j>i.
\end{cases}\\[0.45em]
\end{gathered}
\qquad
\begin{gathered}
R=
\begin{bmatrix}
S_1 & -\infty & -\infty & \cdots\\
E_{21}-\lambda_\ell\log 2 & S_2 & -\infty & \cdots\\
E_{31}-\lambda_\ell\log 3 & E_{32}-\lambda_\ell\log 3 & S_3 & \cdots\\
\vdots & \vdots & \vdots & \ddots
\end{bmatrix}.
\end{gathered}
\end{gathered}
\label{eq:causal-relation}
\end{equation}

The Self temperature \(\tau_S>0\) controls the scale and smoothness of the bounded Self mapping. We set \(\tau_S=2\) in all formal experiments.

Aggregate Exchange evidence grows with the number of historical candidates, even without stronger individual Exchange relations. We therefore apply a learnable layer-wise correction \(-\lambda_\ell\log i\) to all historical Exchange entries, following the broader use of logarithmic length corrections in attention mechanisms \citep{ramapuram2024sigmoid}.

Normalized information flow is derived only after the Relation has been constructed and is applied to the information states,

\begin{equation}
\begin{gathered}
F_i=\operatorname{Softmax}\left(R_i\right),\qquad
Y_i=\sum_{j\leq i}F_{ij}I_j,\qquad
X_{\ell+1}=X_\ell+YW_O.
\end{gathered}
\label{eq:relation-flow}
\end{equation}

\subsection{Multi-Head Relation}

We extend Relation to \(H\) heads, each of width \(d_h=d/H\). Each head independently constructs its own relation matrix \(R^{(h)}\) and normalized flow \(F^{(h)}\).

However, the projected information state \(I\) is simply split into \(H\) head-wise slices \(I^{(h)}\). Because each \(R^{(h)}\) is formed independently while \(I^{(h)}\) comes from direct slicing, their head-wise correspondence is imposed. We therefore mix adjacent information slices using learnable Givens rotations \citep{jing2017eunn}. The pairing pattern alternates between successive layers,

\begin{equation}
\begin{gathered}
\begin{bmatrix}
\widetilde I^{(a)}\\
\widetilde I^{(b)}
\end{bmatrix}
=
\begin{bmatrix}
\cos\theta_{\ell,ab} & -\sin\theta_{\ell,ab}\\
\sin\theta_{\ell,ab} & \cos\theta_{\ell,ab}
\end{bmatrix}
\begin{bmatrix}
I^{(a)}\\
I^{(b)}
\end{bmatrix},\\[0.2em]
Y^{(h)}=F^{(h)}\widetilde I^{(h)},\qquad
X_{\ell+1}=X_\ell+\operatorname{Concat}_{h=1}^{H}\!\left(Y^{(h)}\right)W_O.
\end{gathered}
\label{eq:givens-rotation}
\end{equation}

\section{Why Relation}

We now examine what follows from placing Relation before Flow: how Self and Exchange organize Flow, and what determines the resulting allocation.

\subsection{Relation Is Constructed Before Flow}

Attention normalizes pairwise compatibility scores directly into token-to-token flow. Relation does not. Pairwise evidence is first organized into Self and Exchange and assembled into a role-structured matrix \(R\). Flow is derived only afterward. Relation therefore separates relation construction from flow allocation.

Making \(R\) explicit in the operator does not require materializing the full matrix in memory. Its entries can be constructed and consumed on demand.
\subsection{How Relation Forms Self--Exchange Flow}

Because Self and Exchange are explicit in \(R\), Flow inherits an exact two-level structure. For token \(i>1\), define the historical Exchange log-normalizer \(L_i^E\), its count-corrected aggregate \(A_i\), and the full row normalizer \(Z_i\) as

\begin{equation}
\begin{gathered}
\operatorname{LSE}_{j<i}(a_j)
\equiv\log\sum_{j<i}e^{a_j},
\qquad
L_i^E=\operatorname{LSE}_{j<i}(E_{ij}),\\
A_i
=L_i^E-\lambda_\ell\log i=\operatorname{LSE}_{j<i}(R_{ij}),
\qquad
Z_i=e^{S_i}+e^{A_i}.
\end{gathered}
\label{eq:aggregate-exchange-logit}
\end{equation}

The first level divides Flow between Self and aggregate Exchange,

\begin{equation}
\begin{gathered}
F_{ii}
=
\frac{e^{S_i}}{Z_i}
=
\frac{e^{S_i}}{e^{S_i}+e^{A_i}}
=
\sigma(S_i-A_i),
\qquad
F_{ij}
=
\frac{e^{R_{ij}}}{Z_i},
\quad j<i,\\
g_i
\equiv
\sum_{j<i}F_{ij}
=
\frac{\sum_{j<i}e^{R_{ij}}}{Z_i}
=
\frac{e^{A_i}}{Z_i}
=
\sigma(A_i-S_i),
\qquad
F_{ii}+g_i=1,\\
\begin{aligned}
Y_i
&=
F_{ii}\widetilde I_i
+
\sum_{j<i}F_{ij}\widetilde I_j
=
(1-g_i)\widetilde I_i
+
\sum_{j<i}F_{ij}\widetilde I_j.
\end{aligned}
\end{gathered}
\label{eq:self-flow-mass}
\end{equation}

Thus, \(F_{ii}\) and \(g_i\) form an exact Self--Exchange split. Conditional on Exchange, the second level allocates the Exchange mass across historical tokens according to

\begin{equation}
\begin{gathered}
\begin{aligned}
\pi^E_{ij}
&\equiv
\frac{F_{ij}}{g_i}
=
\frac{e^{R_{ij}}/Z_i}
{\left(\sum_{k<i}e^{R_{ik}}\right)/Z_i}
=
\frac{e^{R_{ij}}}
{\sum_{k<i}e^{R_{ik}}}
\\
&=
\frac{e^{E_{ij}-\lambda_\ell\log i}}
{\sum_{k<i}e^{E_{ik}-\lambda_\ell\log i}}
=
\frac{e^{E_{ij}}}
{\sum_{k<i}e^{E_{ik}}},
\qquad j<i,\\
\end{aligned}\\[0.2em]
\sum_{j<i}\pi^E_{ij}
=1,
\qquad
F_{ij}=g_i\pi^E_{ij},
\qquad
Y_i=(1-g_i)\widetilde I_i
+g_i\sum_{j<i}\pi^E_{ij}\widetilde I_j.
\end{gathered}
\label{eq:conditional-exchange-distribution}
\end{equation}

The same two-level structure has a direct interpretation in information space. Define the Exchange aggregate and its displacement from Self as

\begin{equation}
\begin{aligned}
\bar I_i^E
&\equiv
\sum_{j<i}\pi^E_{ij}\widetilde I_j,
\qquad
\Delta_i^E
\equiv
\bar I_i^E-\widetilde I_i,\\
Y_i
&=
(1-g_i)\widetilde I_i
+
g_i\bar I_i^E
=
\widetilde I_i
+
g_i\Delta_i^E.
\end{aligned}
\label{eq:information-space-form}
\end{equation}

Here, \(\Delta_i^E\) points from the Self state toward aggregate Exchange, while \(g_i\) determines how far the output moves in that direction. As \(g_i\) increases from 0 to 1, \(Y_i\) moves continuously from \(\widetilde I_i\) toward \(\bar I_i^E\). Since \(0\le g_i\le1\), \(Y_i\) lies exactly on the segment between them.

The normalized flow therefore has two levels. It first determines how much mass remains with Self and how much enters Exchange. Conditional on Exchange, it then determines how that mass is distributed across historical tokens. Intuitively, Relation first asks whether a token should rely on itself or draw from its history, and if it draws from history, it then decides where to look.
\subsection{Inside Relation: How a Token's Flow Is Determined}

For an individual token \(i\), the first level is determined by the gap between aggregate Exchange and Self:

\begin{equation}
\log\frac{g_i}{F_{ii}}
=A_i-S_i
=\operatorname{LSE}_{j<i}\left(E_{ij}\right)-S_i-\lambda_\ell\log i.
\label{eq:self-history-odds}
\end{equation}

A larger \(A_i-S_i\) shifts Flow toward Exchange, while a smaller \(A_i-S_i\) shifts it toward Self.

Conditional on Exchange, the second level determines allocation within history. For any two historical positions \(j,k<i\),

\begin{equation}
R_{ij}-R_{ik}=E_{ij}-E_{ik}.
\label{eq:within-history-preference}
\end{equation}

The row-wise correction \(-\lambda_\ell\log i\) cancels in these pairwise differences. It therefore changes the Self--Exchange split without changing allocation within Exchange. That allocation is determined entirely by the Exchange relations \(E_{ij}\).

The same two-level Relation structure also appears in backward credit assignment: gradients first act on the Self--Exchange balance and then on allocation among historical Others. Appendix~A.1.7 gives the full derivation.

\section{Relation in Practice}

The same Self--Exchange organization gives rise to a family of practical Relation forms. We call the explicit token-to-token operator defined in Sec. 3 Full Relation. FlashRelation is its exact fused execution, while Linear Relation represents historical Relation through a recurrent state. Hybrid Relation composes Full and Linear layers within a single decoder.

\subsection{Factorized Self--Exchange: FlashRelation}

The Relation factorization derived in Sec. 4.2 separates historical Exchange aggregation from the final allocation between Self and Exchange. This factorization leads us to develop FlashRelation, a tiled realization of Full Relation. For token \(i\), historical Exchange entries are processed through a causal tiled scan. For each tile \(\mathcal B\), we maintain a running maximum \(m\), normalizer \(l\), and information accumulator \(z\):

\begin{equation}
\begin{gathered}
m'=\max\left(
 m,\max_{j\in\mathcal B}E_{ij}
\right),\\
l'=
e^{m-m'}l
 +
\sum_{j\in\mathcal B}e^{E_{ij}-m'},
\qquad
z'=
e^{m-m'}z
 +
\sum_{j\in\mathcal B}e^{E_{ij}-m'}\widetilde I_j .
\end{gathered}
\label{eq:flashrelation-exchange-scan}
\end{equation}

After each tile, \((m,l,z)\leftarrow(m',l',z')\). Let \((m_i,l_i,z_i)\) denote the terminal accumulators after the historical Exchange scan. They yield

\begin{equation}
L_i^E=m_i+\log l_i,
\qquad
\bar I_i^E=\frac{z_i}{l_i},
\label{eq:flashrelation-exchange-summary}
\end{equation}

where \(L_i^E\) is the aggregate historical Exchange log-normalizer and \(\bar I_i^E\) is the normalized historical information summary. The row is then completed by

\begin{equation}
A_i=L_i^E-\lambda_\ell\log i,
\qquad
g_i=\sigma(A_i-S_i),
\qquad
Y_i=(1-g_i)\widetilde I_i+g_i\bar I_i^E .
\label{eq:flashrelation-factorized-output}
\end{equation}

Following the online-softmax tiling principle of FlashAttention \citep{dao2022flashattention}, the historical Exchange scan is evaluated tile by tile without materializing the full \(R\) or \(F\).

\subsection{Change Exchange: Linear and Hybrid Relation}

Linear Relation preserves Self and changes Exchange. Self remains tied to the current token, while Exchange changes from explicit interaction with individual historical tokens to a recurrent relational state. Historical Others are compressed into
\[
C_t^{(h)} \in \mathbb{R}^{d_h\times d_h}.
\]
The Self relation keeps the construction of Sec. 3.1. Using the same projections, Linear Relation computes \(S_t\) together with two normalized relation coordinates as

\begin{equation}
\begin{gathered}
S_t=\sigma\left(
\frac{p_{1,t}^{\top}p_{2,t}}
{\tau_S\sqrt{d_h}}
\right),\\
\hat p_{1,t}=\frac{p_{1,t}}{\|p_{1,t}\|_2},
\qquad
\hat p_{2,t}=\frac{p_{2,t}}{\|p_{2,t}\|_2}.
\end{gathered}
\label{eq:linear-self-relation}
\end{equation}

The recurrent state carries historical Others through an input-dependent channel-wise retention. The same RoPE used in Sec. 3.1 is applied to \(p_{1,t}\) and \(p_{2,t}\) before \(S_t\) and the normalized coordinates are formed. We parameterize this retention following KDA \citep{kimiLinear2025} as

\begin{equation}
\begin{gathered}
r_t=W_\alpha^\uparrow W_\alpha^\downarrow \bar x_t,
\qquad
g_t=-\exp(A_{\log})\odot\operatorname{softplus}(r_t+dt_{\mathrm{bias}}),
\qquad
\alpha_t=\exp(g_t)\in(0,1]^{d_h}.
\end{gathered}
\label{eq:retention-vector}
\end{equation}

With historical Others compressed into a recurrent state, Self remains explicit in both reading history and writing the current token into that state. Linear Relation therefore follows

\begin{equation}
\begin{gathered}
\boxed{
\text{Ask Self}
\rightarrow
\text{Ask Others}
\rightarrow
\text{Answer}
\rightarrow
\text{Become Others}
}
\\[0.4em]
\begin{aligned}
\text{Ask Self:}\quad
&S_t=
\sigma\!\left(
\frac{
p_{1,t}^{\top}p_{2,t}
}{
\tau_S\sqrt{d_h}
}
\right),\\
\text{Ask Others:}\quad
&E_t=
S_t
\left(
\operatorname{Diag}(\alpha_t)C_{t-1}
\right)^{\top}
\hat p_{1,t},\\
\text{Answer:}\quad
&Y_t=\widetilde I_t+E_t,\\
\text{Become Others:}\quad
&C_t=
\operatorname{Diag}(\alpha_t)C_{t-1}
+S_t\hat p_{2,t}\widetilde I_t^\top.
\end{aligned}
\end{gathered}
\label{eq:decayed-relation-state}
\end{equation}

Unrolling the recurrent state exposes how each historical Other contributes to the current Exchange:

\[
E_t
=
\sum_{j<t}
\underbrace{S_t}_{\text{read by Self}}
\underbrace{S_j}_{\text{written by Self}}
\left[
\hat p_{2,j}^{\top}
\underbrace{
\left(
\prod_{u=j+1}^{t} D_u
\right)
}_{\text{retention}}
\hat p_{1,t}
\right]
\underbrace{\widetilde I_j}_{\text{historical Other}},
\qquad
D_u=\operatorname{Diag}(\alpha_u).
\]

Self therefore remains active on both sides of recurrent Exchange: \(S_j\) when a token becomes historical, and \(S_t\) when history is read. Linear Relation changes how Exchange is represented while preserving the Self--Exchange organization. The full derivation is given in Appendix~A.4.

Exchange remains strictly historical because the retained state is read before the current token is written into it.

Changing Exchange to a recurrent state also changes its computational regime. Table 1 summarizes the resulting token-mixing complexity and decode-state requirements.

\begin{table}[!ht]
\caption{Complexity refers to the token-mixing core. Linear sequential operations refer to the direct recurrence.}
\label{tab:token-mixing-complexity}
\centering
\small
\setlength{\tabcolsep}{3pt}
\begin{tabular}{@{}l c c c@{}}
\toprule
Layer Type & \begin{tabular}[t]{@{}c@{}}Token-Mixing\\Complexity\end{tabular} & Sequential Ops. & \begin{tabular}[t]{@{}c@{}}Decode History\\State\end{tabular} \\
\midrule
MHA / FlashAttention & \(O(T^2 d)\) & \(O(1)\) & \(O(Td)\) \\
Full / FlashRelation & \(O(T^2 d)\) & \(O(1)\) & \(O(Td)\) \\
Linear Relation & \(\mathbf{O(Td^2/H)}\) & \(O(T)\) & \(\mathbf{O(d^2/H)}\) \\
\bottomrule
\end{tabular}
\end{table}

Full and Linear Relation occupy complementary execution regimes. Linear Relation scales linearly with sequence length and uses a fixed-size decode state, while Full Relation retains \(O(1)\) sequential operations. Hybrid Relation combines both within a single decoder by interleaving Linear and Full Relation layers. Following the hybrid layerwise design used in Kimi Linear \citep{kimiLinear2025}, our evaluated configuration uses nine Linear Relation layers and three Full Relation layers.

\section{Experiments}

We compare MHA and Full Relation under matched decoder geometry, data order, and training budget. Experiments cover approximately 10M, 30M, and 100M parameters and use paired seeds 42, 43, and 44. Scale-specific architectures, datasets, context lengths, and token budgets are summarized in Table 2. Our primary metric is final-checkpoint NLL on the full validation set, and structural ablations use the 10M setting with the same three seeds. We additionally report matched-train-loss token reduction, measuring how many fewer training tokens Relation requires to reach the paired MHA run's final smoothed training loss. All models use full-head RoPE with base \(10^4\) and no positional scaling. Systems benchmarks are conducted separately on an NVIDIA RTX 5090 in BF16 and report tuned steady-state throughput. Full optimization, evaluation, and per-seed details are provided in Appendices B and C.

\subsection{Language Modeling and Structural Ablations}

Full Relation improves mean final-validation NLL at all three model scales while also reaching the paired MHA final training loss with fewer tokens (Table 2). The corresponding token reductions are 7.3\%, 5.7\%, and 4.5\% at 10M, 30M, and 100M.

\begin{table}[!ht]
\caption{Main validation results. NLL is mean $\pm$ sample standard deviation over three seeds. Backbone reports layers / $d_{\mathrm{model}}$ / heads $\times d_h$ / $d_{\mathrm{ff}}$. Token reduction reports fewer Relation training tokens required to reach the paired MHA run's final smoothed train loss.}
\label{tab:main-results}
\centering
\scriptsize
\setlength{\tabcolsep}{1.5pt}
\begin{tabular*}{\linewidth}{@{\extracolsep{\fill}}l l l r r r r r r@{}}
\toprule
Scale & Backbone & Data/Vocab & Ctx & Tokens & MHA NLL $\downarrow$ & Relation NLL $\downarrow$ & $\Delta$ & \begin{tabular}[t]{@{}c@{}}Token\\reduction $\uparrow$\end{tabular} \\
\midrule
10M & 6L/384/8$\times$48/768 & TinyStories/4K & 1024 & 150M & $1.6853 \pm 0.0042$ & $\mathbf{1.6441 \pm 0.0124}$ & $-0.0412$ & $7.3 \pm 3.4\%$ \\
30M & 10L/512/8$\times$64/1024 & TinyStories/4K & 2048 & 450M & $1.3001 \pm 0.0044$ & $\mathbf{1.2850 \pm 0.0136}$ & $-0.0151$ & $5.7 \pm 4.9\%$ \\
100M & 20L/640/8$\times$80/1280 & SmolLM/32K & 4096 & 1.071B & $2.9373 \pm 0.0093$ & $\mathbf{2.9063 \pm 0.0061}$ & $-0.0310$ & $4.5 \pm 3.4\%$ \\
\bottomrule
\end{tabular*}
\end{table}
\FloatBarrier

The Full Relation structure matters. Exchange-only transport and Raw-\(X\) communication increase NLL by 0.0320 and 0.0366, while removing count calibration or collapsing Relation to a single head produces larger degradations of about 0.051 (Table 3). The exact ablation definitions and per-seed results are provided in Appendix D.

\begin{table}[!ht]
\caption{Structural ablations on the 10M setting. NLL is mean $\pm$ sample standard deviation over three seeds.}
\label{tab:structural-ablations}
\centering
\small
\setlength{\tabcolsep}{2pt}
\resizebox{\linewidth}{!}{%
\begin{tabular}{@{}l r r r r r@{}}
\toprule
Metric & \begin{tabular}[t]{@{}c@{}}Full Relation\\\end{tabular} & \begin{tabular}[t]{@{}c@{}}Exchange-only\\transport\\$Y_i=\sum_{j<i}F_{ij}\widetilde I_j$\end{tabular} & \begin{tabular}[t]{@{}c@{}}Raw-$X$\\communication\\$I=X$\end{tabular} & \begin{tabular}[t]{@{}c@{}}No count\\calibration\\$\lambda_\ell=0$\end{tabular} & \begin{tabular}[t]{@{}c@{}}Single head\\$H=1$\end{tabular} \\
\midrule
NLL $\downarrow$ & $\mathbf{1.6441 \pm 0.0124}$ & $1.6761 \pm 0.0097$ & $1.6807 \pm 0.0074$ & $1.6947 \pm 0.0203$ & $1.6948 \pm 0.0221$ \\
$\Delta$ vs Full & \textemdash & $+0.0320$ & $+0.0366$ & $+0.0506$ & $+0.0507$ \\
\bottomrule
\end{tabular}
}
\end{table}
\FloatBarrier

\subsection{Is Relation Attention?}

Both Relation and Attention ultimately produce normalized token-to-token flow. Their learned organization, however, is not the same. We compare matched checkpoints at 10M, 30M, and 100M across three paired seeds, focusing on two questions: how flow is organized across depth and how historical flow is routed across tokens (Table~\ref{tab:relation_internal}).

\begin{table}[!ht]
\centering
\caption{Internal organization of matched MHA and Relation checkpoints.
(a) Depth-wise organization of diagonal/Self flow and routing rank.
(b) Conditional historical routing across distance and first-token sink.
Values are three-seed means. Metric definitions and sampling protocol are given in Appendix~\ref{app:structural-diagnostics}.}
\label{tab:relation_internal}
{\scriptsize\emph{Arrows denote MHA $\rightarrow$ Relation.}\par}
\noindent
\begin{minipage}{\linewidth}
\centering
\footnotesize
\setlength{\tabcolsep}{0.5pt}
\renewcommand{\arraystretch}{0.9}
\begin{tabular*}{\linewidth}{@{\extracolsep{\fill}}l c c c c@{}}
\toprule
\multicolumn{5}{c}{\textbf{(a) Depth organization}}\\
Scale &
\begin{tabular}[t]{@{}c@{}}L0 diag. /\\Self flow\end{tabular} &
\begin{tabular}[t]{@{}c@{}}L0 Flow\\EffRank\end{tabular} &
\begin{tabular}[t]{@{}c@{}}L1 diag. /\\Self flow\end{tabular} &
\begin{tabular}[t]{@{}c@{}}Rel. L2+\\Self / Ex.\end{tabular} \\
\midrule
10M & $6.62\%\rightarrow\mathbf{28.57\%}$ & $173.3\rightarrow405.4\;(\mathbf{2.34\times})$ & $7.30\%\rightarrow1.49\%$ & $2.87\%/\mathbf{97.13\%}$\\
30M & $5.54\%\rightarrow\mathbf{23.80\%}$ & $155.4\rightarrow406.9\;(\mathbf{2.62\times})$ & $6.53\%\rightarrow0.37\%$ & $1.73\%/\mathbf{98.27\%}$\\
100M & $13.94\%\rightarrow\mathbf{43.81\%}$ & $176.6\rightarrow470.8\;(\mathbf{2.67\times})$ & $1.86\%\rightarrow0.39\%$ & $1.01\%/\mathbf{98.99\%}$\\
\bottomrule
\end{tabular*}
\end{minipage}
\vspace{2pt}\par\noindent
\begin{minipage}{\linewidth}
\centering
\footnotesize
\setlength{\tabcolsep}{0.5pt}
\renewcommand{\arraystretch}{0.9}
\begin{tabular*}{\linewidth}{@{\extracolsep{\fill}}l c c c c@{}}
\toprule
\multicolumn{5}{c}{\textbf{(b) Historical routing}}\\
Scale & $d>128$ & $d>256$ & $d>512$ &
\begin{tabular}[t]{@{}c@{}}First-token\\sink\end{tabular} \\
\midrule
10M & $18.12\%\rightarrow25.21\%$ & $10.43\%\rightarrow15.63\%$ & $4.17\%\rightarrow6.09\%$ & $0.566\%\rightarrow0.483\%$ \\
30M & $18.17\%\rightarrow23.94\%$ & $9.40\%\rightarrow14.00\%$ & $3.18\%\rightarrow5.23\%$ & $0.772\%\rightarrow0.579\%$ \\
100M & $37.99\%\rightarrow39.58\%$ & $22.59\%\rightarrow23.91\%$ & $8.03\%\rightarrow8.51\%$ & $10.17\%\rightarrow8.43\%$ \\
\bottomrule
\end{tabular*}
\end{minipage}
\end{table}
\FloatBarrier

Relation uses relation. At L0, Relation combines substantially higher Self flow than MHA's diagonal flow with a 2.34--2.67$\times$ higher flow effective rank. At L1, Self flow falls below 1.5\% at every scale. From L2 onward, 97.13--98.99\% of Relation's normalized flow is assigned to Exchange. Unlike MHA, Relation uses L0 as a Self anchor and a high-rank router, then rapidly shifts into an Exchange-dominant regime.
Historical routing also shifts outward. Conditional historical mass beyond 128, 256, and 512 tokens is higher at every scale, while the first-token sink is lower at every scale.

Relation is not Attention. Across all three scales, it learns a distinct depth organization. In Attention, the normalized token-to-token flow is the Attention itself. In Relation, Flow is only the consequence of a prior Relation. Attention ends at Flow. Relation begins before Flow.

\subsection{Efficient Implementations}

In a fixed-context \(T=1024\) reference benchmark, FlashRelation is \(4.17\text{--}5.28\times\) faster than the materialized Full Relation reference implementation. Across the three scale-matched production workloads, FlashRelation reaches \(89.7\text{--}92.9\%\) of PyTorch FlashAttention throughput while executing the exact Full Relation operator (Table 5). Linear Relation throughput results are reported in Appendix H.

\begin{table}[!ht]
\centering
\caption{FlashRelation systems results on an RTX 5090 in BF16. (a) Fixed-context (\(T=1024\)) comparison with the materialized Full Relation reference. (b) Scale-matched production throughput against PyTorch FlashAttention.}
\label{tab:efficient-relation}
\noindent
\begin{minipage}[t]{0.43\linewidth}
\centering
\footnotesize
\setlength{\tabcolsep}{1.5pt}
\begin{tabular}{@{}lrrr@{}}
\toprule
\multicolumn{4}{c}{\textbf{(a) Fixed-context \(T=1024\)}}\\
Scale & Ref. Full & FlashRelation & Speedup \\
\midrule
10M  & 128,515 & 679,121 & $\mathbf{5.28\times}$ \\
30M  & 74,376  & 365,766 & $\mathbf{4.92\times}$ \\
100M & 30,631  & 127,715 & $\mathbf{4.17\times}$ \\
\bottomrule
\end{tabular}
\end{minipage}\hfill
\begin{minipage}[t]{0.55\linewidth}
\centering
\footnotesize
\setlength{\tabcolsep}{1.2pt}
\begin{tabular}{@{}lrrrr@{}}
\toprule
\multicolumn{5}{c}{\textbf{(b) Scale-matched production}}\\
Scale & $T$ & FlashAttn. & FlashRelation & FR / FA \\
\midrule
10M  & 1024 & 841,834 & 755,088 & $\mathbf{0.897\times}$ \\
30M  & 2048 & 361,793 & 334,028 & $\mathbf{0.923\times}$ \\
100M & 4096 & 107,783 & 100,167 & $\mathbf{0.929\times}$ \\
\bottomrule
\end{tabular}
\end{minipage}
\end{table}

Hybrid Relation combines nine Linear and three Full Relation layers in a \((LLLF)^3\) layout. Under the 30M-class training setting, the 31.97M-parameter model achieves \(1.2780 \pm 0.0050\) final-validation NLL across three seeds. This demonstrates that Linear and Full Relation layers can be composed in a decoder in which nine of twelve token-mixing layers use Linear Relation.

\section{Limitations and Conclusion}

\textbf{Limitations.} Our experiments are limited to decoder-only language models up to approximately 100M parameters and 1.071B training tokens. Full Relation is evaluated systematically across scales, while Linear and Hybrid Relation are studied only in the configurations reported here. We also fix \(\tau_S=2\), sigmoid for Self, and SiLU for Exchange. Other role-specific mappings may behave differently. Larger models, multimodal settings, and post-training remain unexplored.

\textbf{Conclusion.} We introduced Relation, an alternative token-mixing primitive that organizes pairwise evidence into explicit Self and Exchange relations before deriving information flow. Full Relation retains explicit token-wise history. FlashRelation provides exact factorized execution. Linear Relation compresses historical Others into a recurrent state, and Hybrid Relation combines Full and Linear layers.

Structural diagnostics show a repeatable organization distinct from matched MHA. Relation uses L0 as a Self anchor and a high-rank router, then shifts rapidly into an Exchange-dominant regime. Across three model scales, Full Relation achieves lower final-validation NLL than matched MHA. FlashRelation provides practical execution, while Hybrid Relation retains strong language-modeling quality with 75\% Linear Relation layers.

Together, these results support a broader view of token mixing. Token mixing need not be organized directly around information flow: it can begin with Relation.

\bibliography{references}
\bibliographystyle{iclr2027_conference}

\clearpage
\appendix
\numberwithin{equation}{section}

\begingroup
% Appendix-local table tightening; main-text layout remains unchanged.
\setlength{\textfloatsep}{6pt plus 1pt minus 2pt}
\setlength{\floatsep}{4pt plus 1pt minus 2pt}
\setlength{\intextsep}{6pt plus 1pt minus 2pt}
\setlength{\abovecaptionskip}{3pt}
\setlength{\belowcaptionskip}{2pt}
\renewcommand{\arraystretch}{0.92}
\setlength{\tabcolsep}{4pt}

\section{Detailed Operator Derivations}

This appendix records the operator definitions used throughout the paper. Token indices in the count correction are one-based. A head has width \(d_h\), and the information state after input projection and head mixing is denoted by \(\widetilde I\).

\subsection{Self--Exchange Relation}

\subsubsection{Relation construction}

We first write the construction for one head. If the model-wide projection has width \(d\), the following quantities are understood after selecting one head subspace of width \(d_h=d/H\). Thus
\[
P_1,P_2,I\in\mathbb R^{T\times d_h},
\qquad
P_1=XW_1,\quad P_2=XW_2,\quad I=XW_I.
\]
The head-wise \(P_1\) and \(P_2\) entering \(U\) use full-head RoPE, whereas \(I\) remains unrotated. Writing \(p_{1,i}\) and \(p_{2,j}\) for rows of these projected states,
\[
U_{ij}=\frac{p_{1,i}^{\top}p_{2,j}}{\sqrt{d_h}}.
\]
The Self and Exchange entries are
\[
S_i=\sigma\!\left(\frac{U_{ii}}{\tau_S}\right),
\qquad
E_{ij}=\operatorname{SiLU}(U_{ij}),
\qquad
\tau_S>0,\quad \tau_S=2.
\]
Here \(\tau_S\) is the Self temperature used to set the scale and smoothness of the bounded Self mapping. The count-calibration parameter \(\lambda_\ell\) is one unconstrained FP32 scalar per layer, shared across heads and initialized to \(0.5\). Causality is imposed when these role-specific entries are assembled:
\[
R_{ij}=
\begin{cases}
S_i,&j=i,\\
E_{ij}-\lambda_\ell\log i,&j<i,\\
-\infty,&j>i.
\end{cases}
\]

\subsubsection{Exact row normalizer}

The normalized row is \(F_i=\operatorname{Softmax}(R_i)\). For \(i>1\), its normalizer is
\[
\begin{aligned}
Z_i
&=\sum_{k\le i}e^{R_{ik}}\\
&=e^{S_i}+\sum_{j<i}e^{E_{ij}-\lambda_\ell\log i}\\
&=e^{S_i}+e^{-\lambda_\ell\log i}\sum_{j<i}e^{E_{ij}}\\
&=e^{S_i}+i^{-\lambda_\ell}\sum_{j<i}e^{E_{ij}}.
\end{aligned}
\]
Define the historical Exchange log-normalizer and its count-corrected version by
\[
L_i^E=\operatorname{LSE}_{j<i}(E_{ij})
=\log\sum_{j<i}e^{E_{ij}},
\qquad
A_i=L_i^E-\lambda_\ell\log i.
\]
Then
\[
e^{A_i}
=e^{-\lambda_\ell\log i}\sum_{j<i}e^{E_{ij}},
\qquad
\boxed{Z_i=e^{S_i}+e^{A_i}}.
\]

\subsubsection{Self mass and aggregate Exchange mass}

The diagonal Self probability is
\[
\begin{aligned}
F_{ii}
&=\frac{e^{S_i}}{Z_i}
=\frac{e^{S_i}}{e^{S_i}+e^{A_i}}\\
&=\frac{1}{1+e^{A_i-S_i}}
=\boxed{\sigma(S_i-A_i)}.
\end{aligned}
\]
The total historical Exchange mass is
\[
\begin{aligned}
g_i
&=\sum_{j<i}F_{ij}
=\frac{e^{A_i}}{e^{S_i}+e^{A_i}}\\
&=\frac{1}{1+e^{S_i-A_i}}
=\boxed{\sigma(A_i-S_i)}.
\end{aligned}
\]
Consequently,
\[
\boxed{F_{ii}+g_i=1}.
\]
The Self--Exchange odds are
\[
\frac{g_i}{F_{ii}}
=\frac{e^{A_i}}{e^{S_i}}
=e^{A_i-S_i},
\]
and therefore
\[
\boxed{
\log\frac{g_i}{F_{ii}}
=A_i-S_i
=L_i^E-S_i-\lambda_\ell\log i.
}
\]

\subsubsection{Conditional Exchange allocation}

For a historical position \(j<i\), define its conditional Exchange allocation as
\[
\pi_{ij}^E=\frac{F_{ij}}{g_i}.
\]
Substituting the normalized row and the aggregate Exchange mass gives
\[
\begin{aligned}
\pi_{ij}^E
&=\frac{e^{E_{ij}-\lambda_\ell\log i}/Z_i}
{e^{A_i}/Z_i}\\
&=\frac{e^{E_{ij}-\lambda_\ell\log i}}
{\displaystyle\sum_{k<i}e^{E_{ik}-\lambda_\ell\log i}}\\
&=\frac{e^{E_{ij}}}{\displaystyle\sum_{k<i}e^{E_{ik}}}.
\end{aligned}
\]
The row-level count correction cancels because it is identical for every historical position. Hence
\[
\boxed{
\pi_{ij}^E
=\operatorname{Softmax}_{j<i}(E_{ij})
},
\qquad
\sum_{j<i}\pi_{ij}^E=1,
\]
and the original historical flow entries factor as
\[
\boxed{F_{ij}=g_i\pi_{ij}^E}.
\]

\subsubsection{Count correction}

For any two historical positions \(j,k<i\),
\[
\begin{aligned}
R_{ij}-R_{ik}
&=(E_{ij}-\lambda_\ell\log i)
-(E_{ik}-\lambda_\ell\log i)\\
&=\boxed{E_{ij}-E_{ik}}.
\end{aligned}
\]
Equivalently,
\[
\frac{\pi_{ij}^E}{\pi_{ik}^E}=e^{E_{ij}-E_{ik}}.
\]
Thus \(-\lambda_\ell\log i\) is a row-level Exchange translation. It changes the Self--Exchange mass competition through \(A_i-S_i\), but it does not change conditional Exchange allocation.

\subsubsection{Exact information transport}

Define the normalized historical information summary
\[
\bar I_i^E=\sum_{j<i}\pi_{ij}^E\widetilde I_j.
\]
The original Full Relation transport is
\[
\begin{aligned}
Y_i
&=\sum_{j\le i}F_{ij}\widetilde I_j\\
&=F_{ii}\widetilde I_i+\sum_{j<i}F_{ij}\widetilde I_j\\
&=F_{ii}\widetilde I_i
+g_i\sum_{j<i}\pi_{ij}^E\widetilde I_j\\
&=\boxed{(1-g_i)\widetilde I_i+g_i\bar I_i^E}.
\end{aligned}
\]
Define the Self-to-Exchange displacement by
\[
\Delta_i^E
\equiv
\bar I_i^E-\widetilde I_i.
\]
Therefore,
\[
Y_i
=
\widetilde I_i
+
g_i\Delta_i^E.
\]
Since \(0\le g_i\le1\), this is an exact interpolation between the current Self information state and the conditional Exchange aggregate. Equivalently, \(\Delta_i^E\) specifies the Self-to-Exchange displacement and \(g_i\) specifies its magnitude.

\subsubsection{Backward credit decomposition}

Define:
\[
G_i
\equiv
\frac{\partial\mathcal L}{\partial Y_i}.
\]
From
\[
Y_i
=
\sum_{j\le i}F_{ij}\widetilde I_j
\]
we obtain
\[
\frac{\partial\mathcal L}
{\partial\widetilde I_j}
=
\sum_{i\ge j}F_{ij}G_i.
\]
Using the Self--Exchange factorization,
\[
\frac{\partial\mathcal L}
{\partial\widetilde I_j}
=
(1-g_j)G_j
+
\sum_{i>j}g_i\pi^E_{ij}G_i.
\]

Using
\[
Y_i
=
\widetilde I_i
+
g_i\Delta_i^E
\]
define
\[
G_i
\equiv
\frac{\partial\mathcal L}{\partial Y_i},
\qquad
c_i
\equiv
G_i^\top\Delta_i^E.
\]

\[
\frac{\partial\mathcal L}{\partial g_i}
=
\left(
\frac{\partial\mathcal L}{\partial Y_i}
\right)^\top
\frac{\partial Y_i}{\partial g_i}
=
G_i^\top\Delta_i^E
=
c_i,
\qquad
\frac{\partial Y_i}{\partial g_i}
=
\frac{\partial}{\partial g_i}
\left(
\widetilde I_i+g_i\Delta_i^E
\right)
=
\Delta_i^E.
\]

\[
g_i=\sigma(A_i-S_i),
\]

\[
\frac{\partial g_i}{\partial(A_i-S_i)}
=
\sigma(A_i-S_i)
\left[
1-\sigma(A_i-S_i)
\right]
=
g_i(1-g_i).
\]

\[
\frac{\partial\mathcal L}{\partial(A_i-S_i)}
=
\frac{\partial\mathcal L}{\partial g_i}
\frac{\partial g_i}{\partial(A_i-S_i)}
=
g_i(1-g_i)c_i.
\]

\[
\frac{\partial\mathcal L}{\partial A_i}
=
g_i(1-g_i)c_i,
\qquad
\frac{\partial\mathcal L}{\partial S_i}
=
-g_i(1-g_i)c_i.
\]

Since
\[
A_i
=
\operatorname{LSE}_{j<i}(E_{ij})
-
\lambda_\ell\log i,
\]
\[
\frac{\partial A_i}{\partial E_{ij}}
=
\pi^E_{ij}.
\]
Moreover,
\[
\bar I_i^E
=
\sum_{k<i}\pi^E_{ik}\widetilde I_k
\]
satisfies
\[
\frac{\partial\bar I_i^E}{\partial E_{ij}}
=
\pi^E_{ij}
\left(
\widetilde I_j-\bar I_i^E
\right).
\]
Therefore,
\[
\frac{\partial\mathcal L}{\partial E_{ij}}
=
g_i(1-g_i)c_i\pi^E_{ij}
+
g_i\pi^E_{ij}
G_i^\top
\left(
\widetilde I_j-\bar I_i^E
\right),
\]
which can be written as
\[
\frac{\partial\mathcal L}{\partial E_{ij}}
=
g_i\pi^E_{ij}
\left[
(1-g_i)c_i
+
G_i^\top
\left(
\widetilde I_j-\bar I_i^E
\right)
\right].
\]
The first term propagates credit through the Self--Exchange amount \(g_i\), while the second propagates credit through the conditional Exchange allocation \(\pi_{ij}^E\).

Backward propagation mirrors the same two-stage Relation semantics as the forward pass: first how much flow remains with Self versus enters Exchange, then how that Exchange flow is allocated among historical Others.

For the general Relation transport, let
\[
H_{ij}
\equiv
G_i^\top\widetilde I_j.
\]
Then
\[
\frac{\partial\mathcal L}{\partial F_{ij}}
=
H_{ij}.
\]
Because \(F_i=\operatorname{Softmax}(R_i)\),
\[
\frac{\partial\mathcal L}{\partial R_{ij}}
=
F_{ij}
\left(
H_{ij}
-
\sum_{k\le i}F_{ik}H_{ik}
\right).
\]

For the role mappings, Self gives
\[
\frac{\partial S_i}{\partial U_{ii}}
=
\frac{1}{\tau_S}S_i(1-S_i).
\]
If
\[
E_{ij}=\operatorname{SiLU}(U_{ij}),
\]
then
\[
\frac{\partial E_{ij}}{\partial U_{ij}}
=
\sigma(U_{ij})
+
U_{ij}\sigma(U_{ij})
\bigl(1-\sigma(U_{ij})\bigr).
\]
For
\[
U
=
\frac{P_1P_2^\top}{\sqrt{d_h}},
\]
\[
\nabla_{P_1}\mathcal L
=
\frac{\nabla_U\mathcal L\,P_2}{\sqrt{d_h}},
\qquad
\nabla_{P_2}\mathcal L
=
\frac{\nabla_U^\top\mathcal L\,P_1}{\sqrt{d_h}}.
\]
Finally,
\[
\frac{\partial\mathcal L}{\partial\lambda_\ell}
=
-\sum_i g_i(1-g_i)c_i\log i,
\]
where the sum ranges over the valid \(i>1\) rows in the layer.

\subsubsection{Boundary case}

For the first token, the historical set is empty. Define
\[
\operatorname{LSE}_{j<1}=-\infty,
\qquad
A_1=-\infty.
\]
It follows directly that
\[
\boxed{g_1=0,\qquad F_{11}=1,\qquad Y_1=\widetilde I_1}.
\]

\subsection{Multi--Head Relation}

\subsubsection{Headwise construction}

Let \(d_h=d/H\). After projection and head splitting,
\[
P_1^{(h)},P_2^{(h)},I^{(h)}
\in\mathbb R^{T\times d_h},
\qquad h=1,\ldots,H.
\]
Each head independently applies the Self--Exchange construction of Appendix~A.1, yielding
\[
R^{(h)},\qquad F^{(h)},\qquad Y^{(h)}.
\]
Relation construction is head-local and cross-head interaction is applied only to the information branch.

\subsubsection{Givens block}

For an adjacent head pair \((a,b)\), define
\[
G(\theta)=
\begin{bmatrix}
\cos\theta&-\sin\theta\\
\sin\theta&\cos\theta
\end{bmatrix}.
\]
The information states are mixed by
\[
\begin{bmatrix}\widetilde I^{(a)}\\\widetilde I^{(b)}\end{bmatrix}
=G(\theta_{\ell,ab})
\begin{bmatrix}I^{(a)}\\I^{(b)}\end{bmatrix}.
\]
Since
\[
G(\theta)^\top G(\theta)=I_2,
\]
the transformation is orthogonal. Applied to every information channel of the pair,
\[
\left\|\widetilde I^{(a)}\right\|_2^2
+\left\|\widetilde I^{(b)}\right\|_2^2
=\left\|I^{(a)}\right\|_2^2
+\left\|I^{(b)}\right\|_2^2.
\]
The same scalar angle \(\theta_{\ell,ab}\) acts on the \(d_h\) information channels associated with the pair.

\subsubsection{Alternating pairing}

The formal construction uses an even number of heads, and in all formal experiments, \(H=8\). With one-based head indices, the two pairings are
\[
\mathcal M_A=\{(1,2),(3,4),\ldots,(H-1,H)\},
\qquad
\mathcal M_B=\{(2,3),(4,5),\ldots,(H,1)\}.
\]
They alternate across layers:
\[
\mathcal M_A\longrightarrow\mathcal M_B\longrightarrow
\mathcal M_A\longrightarrow\cdots.
\]
For \(H=8\),
\[
\mathcal M_A=\{(1,2),(3,4),(5,6),(7,8)\},
\qquad
\mathcal M_B=\{(2,3),(4,5),(6,7),(8,1)\}.
\]
There are \(H/2=4\) learnable angles per layer. They are stored in FP32 and initialized to zero.

\subsubsection{Multi-head transport}

After the role-specific flow is formed for each head,
\[
Y^{(h)}=F^{(h)}\widetilde I^{(h)}.
\]
The head outputs are concatenated and projected:
\[
Y=\operatorname{Concat}\left(Y^{(1)},\ldots,Y^{(H)}\right),
\qquad
X_{\ell+1}=X_\ell+YW_O.
\]
Thus the Relation construction remains head-local, Givens rotations act on the information branch, and \(W_O\) combines the head outputs.

\subsection{FlashRelation}

\subsubsection{Historical Exchange invariant}

FlashRelation evaluates the exact factorization of Appendix~A.1 without materializing the full \(T\times T\) matrices. For a fixed query row \(i\), let \(\mathcal J\subseteq\{1,\ldots,i-1\}\) be the historical indices already scanned. The historical Exchange state is defined by the invariant
\[
m=\max_{j\in\mathcal J}E_{ij},
\qquad
l=\sum_{j\in\mathcal J}e^{E_{ij}-m},
\qquad
z=\sum_{j\in\mathcal J}e^{E_{ij}-m}\widetilde I_j.
\]
This scan contains Exchange entries only and Self is not included in \((m,l,z)\). For the empty set,
\[
m=-\infty,\qquad l=0,\qquad z=0.
\]

\subsubsection{Adding one tile}

Let \(\mathcal B\) be the next causal tile and set
\[
m_{\mathcal B}=\max_{j\in\mathcal B}E_{ij},
\qquad
m'=\max(m,m_{\mathcal B}).
\]
The previously scanned terms are rescaled into the new maximum:
\[
\begin{aligned}
\sum_{j\in\mathcal J}e^{E_{ij}-m'}
&=e^{m-m'}\sum_{j\in\mathcal J}e^{E_{ij}-m}\\
&=e^{m-m'}l.
\end{aligned}
\]
Consequently, the updated normalizer is
\[
\boxed{
l'=e^{m-m'}l+\sum_{j\in\mathcal B}e^{E_{ij}-m'}.
}
\]
The same rescaling gives the information accumulator
\[
\boxed{
z'=e^{m-m'}z
+\sum_{j\in\mathcal B}e^{E_{ij}-m'}\widetilde I_j.
}
\]
Together with \(m'=\max(m,m_{\mathcal B})\), these updates preserve the invariant with \(\mathcal J\leftarrow\mathcal J\cup\mathcal B\).

\subsubsection{Terminal accumulators}

After all historical positions \(j<i\) have been scanned,
\[
m_i=\max_{j<i}E_{ij},
\qquad
l_i=\sum_{j<i}e^{E_{ij}-m_i},
\qquad
z_i=\sum_{j<i}e^{E_{ij}-m_i}\widetilde I_j.
\]
The terminal log-normalizer is
\[
\begin{aligned}
m_i+\log l_i
&=m_i+\log\sum_{j<i}e^{E_{ij}-m_i}\\
&=\log\left(e^{m_i}\sum_{j<i}e^{E_{ij}-m_i}\right)\\
&=\log\sum_{j<i}e^{E_{ij}}\\
&=\boxed{L_i^E}.
\end{aligned}
\]

\subsubsection{Historical information summary}

The terminal accumulator gives
\[
\begin{aligned}
\frac{z_i}{l_i}
&=\frac{\displaystyle\sum_{j<i}e^{E_{ij}-m_i}\widetilde I_j}
{\displaystyle\sum_{k<i}e^{E_{ik}-m_i}}\\
&=\frac{\displaystyle\sum_{j<i}e^{E_{ij}}\widetilde I_j}
{\displaystyle\sum_{k<i}e^{E_{ik}}}\\
&=\sum_{j<i}\pi_{ij}^E\widetilde I_j.
\end{aligned}
\]
The common factor \(e^{-m_i}\) cancels. Therefore
\[
\boxed{
\frac{z_i}{l_i}
=\bar I_i^E.
}
\]

\subsubsection{Recovering the Full Relation row}

The row-level count correction is applied after the historical scan:
\[
A_i=m_i+\log l_i-\lambda_\ell\log i,
\qquad
g_i=\sigma(A_i-S_i).
\]
The final transport is
\[
Y_i=(1-g_i)\widetilde I_i+g_i\frac{z_i}{l_i}.
\]
Using the identities in Appendix~A.1,
\[
\boxed{
Y_i^{\mathrm{FlashRelation}}
=Y_i^{\mathrm{FullRelation}}.
}
\]
So FlashRelation changes the execution of the exact operator.

\subsubsection{Materialization and boundary}

FlashRelation does not materialize full \(T\times T\) \(U\), \(R\), or \(F\) matrices. The required pairwise evidence and Exchange entries are formed and consumed tile by tile. The historical state is maintained as running statistics.

\subsection{Linear Relation}

\subsubsection{Relation coordinates and retention}

In the equations below, \(p_{1,t}\) and \(p_{2,t}\) denote the projected relation coordinates after the formal RoPE step. The positional order is
\[
P_1/P_2\longrightarrow\mathrm{RoPE}
\longrightarrow S_t\longrightarrow\mathrm{L2Norm}
\longrightarrow\text{recurrence}.
\]
Thus
\[
S_t=\sigma\!\left(
\frac{p_{1,t}^{\top}p_{2,t}}{\tau_S\sqrt{d_h}}
\right),
\qquad
\hat p_{1,t}=\frac{p_{1,t}}{\|p_{1,t}\|_2},
\qquad
\hat p_{2,t}=\frac{p_{2,t}}{\|p_{2,t}\|_2}.
\]
The retention parameterization is
\[
r_t=W_\alpha^\uparrow W_\alpha^\downarrow\bar x_t,
\qquad
g_t=-\exp(A_{\log})\odot
\operatorname{softplus}(r_t+dt_{\mathrm{bias}}),
\qquad
\alpha_t=\exp(g_t).
\]
Because \(\exp(A_{\log})>0\) elementwise and \(\operatorname{softplus}(\cdot)>0\),
\[
g_t<0,
\qquad
\boxed{0<\alpha_{t,c}\le 1}.
\]
Here \(S_t\) is the Self Relation, whereas \(\alpha_t\) is channel-wise historical retention.

\subsubsection{Recurrent step}

Let
\[
D_t=\operatorname{Diag}(\alpha_t),
\qquad
\boxed{C_0=0}.
\]
The recurrent step is
\[
\begin{aligned}
C_t^-&=D_tC_{t-1},\\
E_t&=S_t(C_t^-)^\top\hat p_{1,t},\\
Y_t&=\widetilde I_t+E_t,\\
C_t&=C_t^-+S_t\hat p_{2,t}\widetilde I_t^\top.
\end{aligned}
\]
The execution order is
\[
\boxed{
\text{Ask Self}
\longrightarrow
\text{Ask Others}
\longrightarrow
\text{Answer}
\longrightarrow
\text{Become Others}.
}
\]

\subsubsection{Strict historical support}

The read uses \(C_t^-=D_tC_{t-1}\), while the current write
\[
S_t\hat p_{2,t}\widetilde I_t^\top
\]
is added only after \(E_t\) has been formed. Hence \(E_t\) depends only on positions \(1,\ldots,t-1\), and the current token cannot enter its own historical read through the current write.

\subsubsection{Unrolling the recurrent state}

Starting from
\[
C_t=D_tC_{t-1}+S_t\hat p_{2,t}\widetilde I_t^\top,
\]
one expansion gives
\[
\begin{aligned}
C_t
&=D_t\left(D_{t-1}C_{t-2}
+S_{t-1}\hat p_{2,t-1}\widetilde I_{t-1}^\top\right)
+S_t\hat p_{2,t}\widetilde I_t^\top\\
&=D_tD_{t-1}C_{t-2}
+D_tS_{t-1}\hat p_{2,t-1}\widetilde I_{t-1}^\top
+S_t\hat p_{2,t}\widetilde I_t^\top.
\end{aligned}
\]
With the ordered product
\[
\prod_{u=j+1}^{t}D_u
\;\equiv\;
D_tD_{t-1}\cdots D_{j+1},
\]
and the empty product equal to the identity,
\[
\boxed{
C_t=\sum_{j\le t}
\left(\prod_{u=j+1}^{t}D_u\right)
S_j\hat p_{2,j}\widetilde I_j^\top.
}
\]
Thus each token contributes \(S_j\hat p_{2,j}\widetilde I_j^\top\) and is subsequently transformed by \(D_{j+1},\ldots,D_t\).

\subsubsection{Expanded historical Exchange}

The pre-write state is
\[
C_t^-=
\sum_{j<t}
\left(\prod_{u=j+1}^{t}D_u\right)
S_j\hat p_{2,j}\widetilde I_j^\top.
\]
Substituting this into the read gives
\[
\begin{aligned}
E_t
&=S_t(C_t^-)^\top\hat p_{1,t}\\
&=\boxed{
\sum_{j<t}
S_tS_j
\left[
\hat p_{2,j}^{\top}
\left(\prod_{u=j+1}^{t}D_u\right)
\hat p_{1,t}
\right]\widetilde I_j
}.
\end{aligned}
\]
Because each \(D_u\) is diagonal, the transpose leaves the ordered product unchanged.

\subsubsection{Information write and complexity}

The output is
\[
Y_t=\widetilde I_t+E_t,
\]
but the state writes
\[
S_t\hat p_{2,t}\widetilde I_t^\top,
\]
not \(S_t\hat p_{2,t}Y_t^\top\). Each new state contribution therefore corresponds to the current token's own information.

For one head, \(C_t^{(h)}\in\mathbb R^{d_h\times d_h}\). The principal recurrent operations per token are
\[
D_tC_{t-1},\qquad
(C_t^-)^\top\hat p_{1,t},\qquad
\hat p_{2,t}\widetilde I_t^\top,
\]
with leading cost \(O(d_h^2)\) per token per head. Across \(H\) heads,
\[
O(THd_h^2)
=O\!\left(TH\left(\frac dH\right)^2\right)
=\boxed{O\!\left(\frac{Td^2}{H}\right)}.
\]
The recurrent state size is
\[
Hd_h^2=H\left(\frac dH\right)^2=\frac{d^2}{H},
\qquad
\boxed{O(d^2/H)}
\]
for the decode state. These complexities refer to the recurrent Relation token-mixing core and exclude common FFN and embedding costs.

\section{Experimental Setup and Evaluation}

\subsection{Model Configurations}

All models use full-head RoPE with base \(10^4\). The three main scales use paired seeds \(42,43,44\).

\begin{table*}[!htbp]
\centering
\caption{Formal decoder-only model configurations. Global tokens per optimizer update are 131,072 for all three scales.}
\label{tab:app-configs}
\scriptsize
\begin{tabular}{@{}l r l r l r l l@{}}
\toprule
Scale & Params MHA/Rel. & L/\(d\)/\(H\times d_h\)/\(d_{\rm ff}\) & Vocab & Data & \(T\) & Train & Micro/GA \\
\midrule

10M & 10,425,216/10,425,246 & 6/384/8\(\times\)48/768 & 4K & TinyStories & 1024 & 150M & 32/4 \\
30M & 28,322,304/28,322,354 & 10/512/8\(\times\)64/1024 & 4K & TinyStories & 2048 & 450M & 16/4 \\
100M & 102,917,760/102,917,860 & 20/640/8\(\times\)80/1280 & 32K & SmolLM corpus & 4096 & 1.071B & MHA 2/16; Rel. 4/8 \\

\bottomrule
\end{tabular}
\end{table*}
\FloatBarrier

\subsection{Optimization and Checkpoints}

Training uses BF16 and AdamW with betas \((0.9,0.95)\), weight decay \(0.1\), gradient clipping at \(1.0\), and a WSD80 schedule. Learning rates are \(10^{-3}\), \(8\times10^{-4}\), and \(6\times10^{-4}\) for 10M, 30M, and 100M. Warmup tokens are 1.5M, 4.5M, and 10.7114M. Formal NLL is evaluated on the full validation set at the final checkpoint.

\subsection{Position Encoding}

RoPE covers the full head dimension. The base is \(10^4\). For MHA, RoPE is applied to \(Q\) and \(K\) after projection and head splitting. For Full Relation and FlashRelation, RoPE is applied to \(P_1\) and \(P_2\) before pairwise evidence. The information projection and Givens output remain unrotated. Linear Relation applies RoPE to its two relation coordinates before Self and normalized coordinate formation. Hybrid layers use the corresponding Full or Linear path.

\subsection{Evaluation Protocol}
\label{app:structural-diagnostics}

The primary metric is final-checkpoint NLL. Systems measurements report complete optimizer-step throughput. One-time initialization and JIT, warmup, and terminal partial steps are excluded when present.

For matched-train-loss token efficiency, training loss is smoothed with a trailing token window equal to 1\% of the formal training budget. For each paired seed, the target is the MHA run's final smoothed training loss. We linearly interpolate the earliest Relation crossing between adjacent logged points and report the token reduction as \(1-N_{\mathrm{Rel}}/N_{\mathrm{budget}}\).

\paragraph{Structural diagnostic definitions.}
MHA$\rightarrow$Rel. reports matched MHA and Relation means. \emph{diag. / Self flow} compares the MHA diagonal normalized-flow entry with the Relation Self-flow entry; the MHA diagonal is not an explicit Self relation, whereas the Relation entry is induced by the explicit Self relation. L2+ Self / Exchange is reported for Relation only and averages normalized Self and Exchange flow over all layers from L2 onward. Historical-distance and first-token-sink statistics are conditioned on historical flow, with first-token sink denoting the conditional mass assigned to the first sequence token. Flow statistics use the same four fixed sequences of length 1024 across matched checkpoints; flow effective rank uses the same four fixed sequences of length 512.

For a per-head normalized flow matrix \(M\), with \(M=A\) for MHA and \(M=F\) for Relation, let \(\sigma_k(M)\) denote singular values computed in float32. The flow effective rank is
\[
\begin{aligned}
\operatorname{EffRank}(M)
&=\exp\!\left(-\sum_k \widetilde p_k\log \widetilde p_k\right),
\\
p_k
&=\frac{\sigma_k(M)}{\max\!\left(\sum_r\sigma_r(M),10^{-20}\right)},
\qquad
\widetilde p_k=\max(p_k,10^{-20}).
\end{aligned}
\]
For each layer and head, the metric pools the four fixed length-512 rank samples. Within each checkpoint, heads are count-weighted at each layer, and scale-level summaries count-weight layers. We report the mean and sample standard deviation across the three seeds.

The remaining flow-spectrum quantities use squared singular values. With \(\sigma_1(M)\ge\sigma_2(M)\ge\cdots\) from the float32 SVD, define
\[
E(M)\equiv\max\!\left(\sum_k\sigma_k^2(M),10^{-20}\right),
\qquad
\operatorname{StableRank}(M)\equiv
\frac{E(M)}{\max\!\left(\sigma_1^2(M),10^{-20}\right)}.
\]
The energy dimensions and top singular energy are
\[
r_\eta(M)\equiv\min\!\left\{r:\frac{\sum_{k=1}^{r}\sigma_k^2(M)}{E(M)}\ge\eta\right\},
\qquad \eta\in\{0.90,0.95\},
\]
\[
\operatorname{TopSingularEnergy}(M)\equiv\frac{\sigma_1^2(M)}{E(M)}.
\]
The 90\% and 95\% dimensions are \(r_{0.90}(M)\) and \(r_{0.95}(M)\), respectively. These spectrum statistics use the same per-head flow matrices, fixed rank samples, sample pooling, head aggregation, checkpoint/layer aggregation, and seed aggregation as Effective Rank.

For conditional sink metrics, we sum \(\pi^E_{ij}\) over zero-indexed source positions \(j<k\) for each query row. Rows without historical Exchange mass, including the first token, contribute zero to the reported mean.

\subsection{Structural Diagnostic Results}
\label{app:structural-diagnostics-results}

The formal structural diagnostics cover all 18 final checkpoints: three scales, three seeds, and matched MHA/Relation pairs. Flow, distance, and sink statistics use the same four fixed sequences of length 1024 for every checkpoint, while spectrum statistics use the same four fixed sequences of length 512. For MHA, the normalized flow is reconstructed from the checkpoint's actual $Q$ and $K$ projections with RoPE, causal masking, and Softmax; Relation uses its actual normalized flow $F$.

For Relation, the aggregate normalized Self/Exchange flow from L2 onward is $2.87\%/97.13\%$ at 10M, $1.73\%/98.27\%$ at 30M, and $1.01\%/98.99\%$ at 100M (three-seed means).

\begin{table*}[!htbp]
\centering
\caption{Layerwise normalized-flow organization in the formal structural diagnostics. MHA diag. is the diagonal normalized-flow entry. Rel. Self is Relation's explicit Self flow. EffRank is the per-head normalized-flow spectrum metric defined in Appendix~B.4. Values are three-seed means.}
\label{tab:app-structural-layerwise}
\scriptsize
\setlength{\tabcolsep}{3.5pt}
\renewcommand{\arraystretch}{0.92}
\begin{tabular}{@{}llrrrrr@{}}
\toprule
Scale & Layer & MHA diag. & Rel. Self & MHA EffRank & Rel. EffRank & Rel./MHA \\
\midrule
10M & L0 & 6.62\% & 28.57\% & 173.3 & 405.4 & 2.34 \\
10M & L1 & 7.30\% & 1.49\% & 244.2 & 216.3 & 0.89 \\
10M & L2 & 4.29\% & 1.87\% & 192.2 & 224.5 & 1.17 \\
10M & L3 & 5.36\% & 3.10\% & 189.0 & 211.1 & 1.12 \\
10M & L4 & 5.76\% & 3.39\% & 173.1 & 211.2 & 1.22 \\
10M & L5 & 6.59\% & 3.13\% & 203.7 & 201.2 & 0.99 \\
\midrule
30M & L0 & 5.54\% & 23.80\% & 155.4 & 406.9 & 2.62 \\
30M & L1 & 6.53\% & 0.37\% & 242.6 & 240.7 & 0.99 \\
30M & L2 & 4.83\% & 1.34\% & 226.3 & 200.2 & 0.88 \\
30M & L3 & 4.43\% & 1.22\% & 197.3 & 200.6 & 1.02 \\
30M & L4 & 4.14\% & 1.27\% & 181.7 & 180.8 & 1.00 \\
30M & L5 & 4.11\% & 1.69\% & 179.2 & 181.5 & 1.01 \\
30M & L6 & 3.85\% & 2.19\% & 169.8 & 201.1 & 1.18 \\
30M & L7 & 4.98\% & 1.65\% & 186.3 & 182.5 & 0.98 \\
30M & L8 & 5.04\% & 2.34\% & 184.3 & 190.4 & 1.03 \\
30M & L9 & 5.21\% & 2.17\% & 189.7 & 218.6 & 1.15 \\
\midrule
100M & L0 & 13.94\% & 43.81\% & 176.6 & 470.8 & 2.67 \\
100M & L1 & 1.86\% & 0.39\% & 129.7 & 185.0 & 1.43 \\
100M & L2 & 3.23\% & 1.19\% & 159.9 & 202.8 & 1.27 \\
100M & L3 & 3.01\% & 1.44\% & 166.2 & 192.4 & 1.16 \\
100M & L4 & 3.76\% & 1.36\% & 174.3 & 187.0 & 1.07 \\
100M & L5 & 4.27\% & 1.65\% & 171.9 & 198.7 & 1.16 \\
100M & L6 & 3.29\% & 2.03\% & 175.0 & 195.8 & 1.12 \\
100M & L7 & 3.08\% & 1.48\% & 136.6 & 179.7 & 1.32 \\
100M & L8 & 2.56\% & 1.12\% & 140.0 & 168.0 & 1.20 \\
100M & L9 & 2.84\% & 2.13\% & 147.8 & 159.7 & 1.08 \\
100M & L10 & 3.52\% & 1.22\% & 152.1 & 155.6 & 1.02 \\
100M & L11 & 2.54\% & 1.02\% & 141.7 & 145.6 & 1.03 \\
100M & L12 & 2.97\% & 0.39\% & 124.6 & 152.4 & 1.22 \\
100M & L13 & 3.43\% & 0.23\% & 164.3 & 146.5 & 0.89 \\
100M & L14 & 3.39\% & 0.33\% & 157.0 & 158.4 & 1.01 \\
100M & L15 & 3.80\% & 0.17\% & 145.3 & 142.9 & 0.98 \\
100M & L16 & 4.37\% & 0.27\% & 165.3 & 141.3 & 0.85 \\
100M & L17 & 4.97\% & 0.69\% & 141.2 & 154.8 & 1.10 \\
100M & L18 & 7.41\% & 0.62\% & 172.1 & 152.7 & 0.89 \\
100M & L19 & 6.06\% & 0.82\% & 153.4 & 148.4 & 0.97 \\
\bottomrule
\end{tabular}
\end{table*}
\FloatBarrier

\begin{table*}[!htbp]
\centering
\caption{Scale-level conditional historical routing in the formal structural diagnostics. Each entry is shown as MHA $\rightarrow$ Relation; distance mass and first-token sink are conditioned on historical flow. Values are three-seed means.}
\label{tab:app-structural-routing}
\small
\setlength{\tabcolsep}{5pt}
\begin{tabular}{@{}lcccc@{}}
\toprule
Scale & $d>128$ & $d>256$ & $d>512$ & First-token sink \\
\midrule
10M & $18.12\%\rightarrow25.21\%$ & $10.43\%\rightarrow15.63\%$ & $4.17\%\rightarrow6.09\%$ & $0.566\%\rightarrow0.483\%$ \\
30M & $18.17\%\rightarrow23.94\%$ & $9.40\%\rightarrow14.00\%$ & $3.18\%\rightarrow5.23\%$ & $0.772\%\rightarrow0.579\%$ \\
100M & $37.99\%\rightarrow39.58\%$ & $22.59\%\rightarrow23.91\%$ & $8.03\%\rightarrow8.51\%$ & $10.17\%\rightarrow8.43\%$ \\
\bottomrule
\end{tabular}
\end{table*}
\FloatBarrier

\begin{table*}[!htbp]
\centering
\caption{Conditional early-token sink profile in the formal structural diagnostics. All sink metrics are conditioned on historical flow. \emph{First} \(k\) denotes the conditional historical-flow mass assigned to the first \(k\) sequence positions. Arrows denote MHA \(\rightarrow\) Relation; values are three-seed means.}
\label{tab:app-structural-sink-profile}
\scriptsize
\setlength{\tabcolsep}{3pt}
\begin{tabular}{@{}c@{}}
\begin{tabular}{@{}lccc@{}}
\toprule
\multicolumn{4}{c}{(a) First 1 / 2 / 4} \\
\midrule
Scale & First 1 & First 2 & First 4 \\
\midrule
10M & $0.566\%\rightarrow0.483\%$ & $0.990\%\rightarrow0.908\%$ & $1.552\%\rightarrow1.534\%$ \\
30M & $0.772\%\rightarrow0.579\%$ & $1.217\%\rightarrow1.055\%$ & $1.763\%\rightarrow1.688\%$ \\
100M & $10.173\%\rightarrow8.430\%$ & $10.427\%\rightarrow8.730\%$ & $10.917\%\rightarrow9.257\%$ \\
\bottomrule
\end{tabular}
\\[0.6em]
\begin{tabular}{@{}lccc@{}}
\toprule
\multicolumn{4}{c}{(b) First 8 / 16 / 32} \\
\midrule
Scale & First 8 & First 16 & First 32 \\
\midrule
10M & $2.563\%\rightarrow2.682\%$ & $4.315\%\rightarrow4.721\%$ & $7.157\%\rightarrow8.037\%$ \\
30M & $2.855\%\rightarrow2.885\%$ & $4.616\%\rightarrow4.872\%$ & $7.325\%\rightarrow8.077\%$ \\
100M & $11.839\%\rightarrow10.270\%$ & $13.576\%\rightarrow12.209\%$ & $17.636\%\rightarrow16.506\%$ \\
\bottomrule
\end{tabular}
\end{tabular}
\end{table*}
\FloatBarrier

The full sink profile shows that Relation's lower first-token sink does not imply uniformly lower mass over the entire early context; instead, early historical mass is redistributed across a broader prefix.

\begin{table*}[!htbp]
\centering
\caption{Scale-level aggregate flow-spectrum statistics in the formal structural diagnostics. Each entry is shown as MHA $\rightarrow$ Relation; values are computed from the normalized flow spectra and are three-seed means.}
\label{tab:app-structural-spectrum}
\scriptsize
\setlength{\tabcolsep}{3pt}
\begin{tabular}{@{}lccccc@{}}
\toprule
Scale & Stable rank & Effective rank & 90\% dims & 95\% dims & Top singular energy \\
\midrule
10M & $23.447\rightarrow24.160$ & $195.924\rightarrow244.945$ & $77.07\rightarrow115.67$ & $101.70\rightarrow147.85$ & $0.07446\rightarrow0.06399$ \\
30M & $18.982\rightarrow19.879$ & $191.268\rightarrow220.353$ & $70.64\rightarrow93.32$ & $95.52\rightarrow123.35$ & $0.08551\rightarrow0.08239$ \\
100M & $8.981\rightarrow11.055$ & $154.758\rightarrow181.916$ & $49.83\rightarrow66.55$ & $69.38\rightarrow89.11$ & $0.41186\rightarrow0.37475$ \\
\bottomrule
\end{tabular}
\end{table*}
\FloatBarrier

Taken together, the diagnostics show a consistent depth pattern. Relation uses L0 as a Self-emphasizing, high-rank routing layer. Self flow then drops sharply, and later layers become Exchange-dominant. Historical routing also shifts farther back in the sequence, especially at 10M and 30M, while the first-token sink is lower. These results show that Relation and MHA organize normalized flow differently across depth and history.

\section{Per-Seed Language-Model Results}

\subsection{Final Validation NLL}

Table~\ref{tab:app-nll} reports final validation NLL for all paired seeds. Aggregate rows use the sample standard deviation over three seeds.

\begin{table*}[!htbp]
\centering
\caption{Final validation NLL for the three paired seeds.}
\label{tab:app-nll}
\small
\begin{tabular}{@{}llrrr@{}}
\toprule
Scale & Seed & MHA NLL & Full Relation NLL & Full--MHA \\
\midrule

10M & 42 & 1.688823 & 1.631738 & -0.057085 \\
10M & 43 & 1.686405 & 1.643978 & -0.042427 \\
10M & 44 & 1.680626 & 1.656633 & -0.023993 \\
10M & Mean \(\pm\) SD & 1.685285 \(\pm\) 0.004212 & 1.644117 \(\pm\) 0.012448 & -0.041168 \\
30M & 42 & 1.295263 & 1.300730 & +0.005467 \\
30M & 43 & 1.303885 & 1.277177 & -0.026708 \\
30M & 44 & 1.301160 & 1.277225 & -0.023935 \\
30M & Mean \(\pm\) SD & 1.300103 \(\pm\) 0.004407 & 1.285044 \(\pm\) 0.013584 & -0.015059 \\
100M & 42 & 2.945050 & 2.912234 & -0.032816 \\
100M & 43 & 2.939826 & 2.899965 & -0.039861 \\
100M & 44 & 2.926899 & 2.906657 & -0.020242 \\
100M & Mean \(\pm\) SD & 2.937258 \(\pm\) 0.009344 & 2.906285 \(\pm\) 0.006143 & -0.030973 \\

\bottomrule
\end{tabular}
\end{table*}
\FloatBarrier

Full Relation wins all three paired comparisons at 10M, two of three at 30M, and all three at 100M, for eight wins out of nine paired comparisons.

\subsection{Matched-Train-Loss Token Efficiency}

\begin{table*}[!htbp]
\centering
\caption{Per-seed matched-train-loss token efficiency. The MHA target is the paired run's final smoothed training loss.}
\label{tab:app-token-efficiency}
\small
\setlength{\tabcolsep}{3pt}
\begin{tabular}{@{}l r r r r@{}}
\toprule
Scale & Seed & MHA target & \begin{tabular}[t]{@{}c@{}}Relation crossing\\tokens\end{tabular} & Token reduction \\
\midrule
10M & 42 & 1.645060 & 143.62M & 4.25\% \\
10M & 43 & 1.696446 & 133.67M & 10.88\% \\
10M & 44 & 1.680547 & 139.89M & 6.74\% \\
30M & 42 & 1.302310 & 419.67M & 6.74\% \\
30M & 43 & 1.256313 & 448.42M & 0.35\% \\
30M & 44 & 1.303514 & 405.50M & 9.89\% \\
100M & 42 & 2.914568 & 1.00168B & 6.48\% \\
100M & 43 & 2.903508 & 1.00316B & 6.35\% \\
100M & 44 & 2.870894 & 1.06466B & 0.60\% \\
\bottomrule
\end{tabular}
\end{table*}
\FloatBarrier

Across the three paired seeds, the mean token reductions are \(7.3 \pm 3.4\%\), \(5.7 \pm 4.9\%\), and \(4.5 \pm 3.4\%\) at 10M, 30M, and 100M, respectively.

\section{Structural Ablation Details}

\subsection{Ablation Definitions}

All structural ablations use the 10M setting, the same prepared data, and seeds 42, 43, and 44. Each ablation changes only the named component of Full Relation.
\begin{itemize}
\item Exchange-only transport computes the complete Full Relation flow and transports only historical entries: \(Y_i=\sum_{j<i}F_{ij}\widetilde I_j\). Historical weights are not renormalized.
\item Raw-\(X\) communication replaces the independent information projection by \(I=X\). Head splitting, Givens mixing, Flow, and \(W_O\) remain unchanged.
\item No count calibration sets \(\lambda_\ell=0\).
\item No Givens mixing bypasses adjacent-head rotation, so \(\widetilde I=I\), while the Relation and transport path remains unchanged.
\item Single head uses \(H=1\).
\end{itemize}

\subsection{Per-Seed Ablation Results}

\begin{table*}[!htbp]
\centering
\caption{Per-seed final validation NLL for the structural ablations.}
\label{tab:app-ablation}
\scriptsize
\begin{tabular}{@{}lrrrrr@{}}
\toprule
Ablation & Seed 42 & Seed 43 & Seed 44 & Mean \(\pm\) SD & \(\Delta\) vs Full \\
\midrule

Full Relation & 1.631738 & 1.643978 & 1.656633 & 1.644117 \(\pm\) 0.012448 & 0.000000 \\
Exchange-only transport & 1.665658 & 1.677815 & 1.684822 & 1.676099 \(\pm\) 0.009697 & +0.031982 \\
Raw-\(X\) communication & 1.672332 & 1.686600 & 1.683090 & 1.680674 \(\pm\) 0.007434 & +0.036557 \\
No count calibration & 1.671373 & 1.704531 & 1.708256 & 1.694720 \(\pm\) 0.020305 & +0.050604 \\
No Givens mixing & 1.630321 & 1.646350 & 1.665386 & 1.647352 \(\pm\) 0.017554 & +0.003236 \\
Single head & 1.669349 & 1.707688 & 1.707422 & 1.694820 \(\pm\) 0.022059 & +0.050703 \\

\bottomrule
\end{tabular}
\end{table*}
\FloatBarrier

\section{Hybrid Relation}

\subsection{Configuration}

Hybrid Relation uses a 12-layer, 480-dimensional decoder with eight heads of width 60 and \(d_{\mathrm{ff}}=960\). Its layout is \((LLLF)^3\), with nine Linear Relation layers and three Full Relation layers. The vocabulary has 4,096 entries, context length is 2,048, and training budget is 450M tokens. The model has 31,967,571 parameters and uses micro/GA \(16/4\), BF16, AdamW, and 131,072 global tokens per update. This is a 30M-class configuration and is not geometry-identical to the 30M Full Relation model.

\subsection{Per-Seed Results}

\begin{table}[!htbp]
\centering
\caption{Per-seed Hybrid Relation results.}
\label{tab:app-hybrid}
\small
\begin{tabular}{@{}lr@{}}
\toprule
Seed & Final NLL \\
\midrule

42 & 1.274777 \\
43 & 1.283791 \\
44 & 1.275435 \\
Mean \(\pm\) SD & 1.278001 \(\pm\) 0.005025 \\

\bottomrule
\end{tabular}
\end{table}
\FloatBarrier

\section{Systems Benchmark Protocol}

\subsection{Hardware and Software}

Systems measurements use an NVIDIA RTX 5090 GPU in BF16. The recorded software environment is CUDA 13.0, PyTorch 2.12.1+cu130, and Triton 3.7.1. Each campaign uses its formal prepared data and frozen model geometry.

\subsection{Timing and Accounting}

One-time initialization and JIT compilation are outside the reported interval. Warmup optimizer steps are excluded. Throughput uses complete optimizer steps, with the terminal partial token step excluded where the campaign reaches a non-integral final update. The denominator is the number of tokens consumed by the complete measured optimizer steps.

\subsection{Campaign Settings}

The fixed-context Reference--FlashRelation campaign uses \(T=1024\) for all scales. The scale-matched FlashRelation--FlashAttention campaign uses \(T=1024,2048,4096\) for 10M, 30M, and 100M. Linear long-context scaling uses micro \(=1\), 131,072 effective tokens per update, five warmup steps, and ten measured complete steps.

\section{FlashRelation Systems Results}

\subsection{Scale-Matched FlashRelation and FlashAttention}

FlashAttention is explicitly forced in these scale-matched production records. FlashRelation uses the final exact production dispatch for each head dimension.

\begin{table*}[!htbp]
\centering
\caption{Scale-matched production FlashAttention and FlashRelation throughput; memory is allocated/reserved GiB.}
\label{tab:app-prod-fr-fa}
\tiny
\begin{tabular}{@{}lrrrrllll@{}}
\toprule
Scale & \(T\) & FlashAttention tok/s & FlashRelation tok/s & FR/FA & FR micro/GA & FA micro/GA & FR alloc./res. & FA alloc./res. \\
\midrule

10M & 1024 & 841,833.930 & 755,088.370 & 0.897000\(\times\) & 32/4 & 32/4 & 5.24/5.70 & 5.38/5.88 \\
30M & 2048 & 361,793.115 & 334,028.010 & 0.923300\(\times\) & 16/4 & 16/4 & 9.68/10.04 & 10.03/10.45 \\
100M & 4096 & 107,783.242 & 100,166.770 & 0.929300\(\times\) & 4/8 & 4/8 & 17.19/18.21 & 17.71/18.63 \\

\bottomrule
\end{tabular}
\end{table*}
\FloatBarrier

\section{Linear Relation Systems and Long-Context Scaling}

\subsection{Steady-State Throughput}

\begin{table}[!htbp]
\centering
\caption{Frozen Linear Relation steady-state throughput.}
\label{tab:app-linear-steady}
\small
\begin{tabular}{@{}lrrrr@{}}
\toprule
Scale & \(T\) & Micro/GA & Steady-state tok/s & Raw mean tok/s \\
\midrule

10M & 1024 & 64/2 & 517,602 & 282,578 \\
30M & 1024 & 32/4 & 256,034 & 199,875 \\
100M & 1024 & 16/8 & 90,592 & 87,865 \\

\bottomrule
\end{tabular}
\end{table}
\FloatBarrier

\subsection{Long-Context Scaling}

Long-context runs use the production \(K_{L60}\) schedule with chunk 128, block 64, four warps, one stage, and row tile 1. Micro is fixed to 1 and the effective update size is 131,072 tokens. The \(T=65536\) point is a real OOM.

\begin{table}[!htbp]
\centering
\caption{Linear Relation long-context scaling.}
\label{tab:app-linear-long}
\small
\begin{tabular}{@{}rrrrrr@{}}
\toprule
\(T\) & Micro & GA & Status & tok/s & Allocated GiB \\
\midrule

1024 & 1 & 128 & PASS & 10,129.47 & 1.140 \\
2048 & 1 & 64 & PASS & 20,301.39 & 1.640 \\
4096 & 1 & 32 & PASS & 39,990.24 & 2.648 \\
8192 & 1 & 16 & PASS & 59,367.95 & 4.819 \\
16384 & 1 & 8 & PASS & 52,894.96 & 8.827 \\
32768 & 1 & 4 & PASS & 30,003.56 & 16.815 \\
65536 & 1 & 2 & OOM & -- & -- \\

\bottomrule
\end{tabular}
\end{table}
\FloatBarrier

\endgroup

\end{document}